\documentclass[10pt]{iopart}

\usepackage{booktabs}
\usepackage{colortbl}
\usepackage{iopams} 
\usepackage{graphicx}
\usepackage{multirow}
\usepackage[justification=raggedright]{caption}
\begin{document}
\newcommand{\beginsupplement}{%
        \setcounter{table}{0}
        \renewcommand{\thetable}{S\arabic{table}}%
        \setcounter{figure}{0}
        \renewcommand{\thefigure}{S\arabic{figure}}%
     }

\title[Machine Learning for Seizure Patient Triage in PICU]{Machine Learning to Support Triage of Children at Risk for Epileptic Seizures in the Pediatric Intensive Care Unit}

\author[1]{Raphael Azriel$^1$, Cecil D. Hahn$^2$, Thomas De Cooman$^3$, Sabine Van Huffel$^3$, Eric T. Payne$^4$, Kristin L. McBain$^5$, Danny Eytan$^6$* and Joachim A. Behar$^1$*}

\address{$^1$ Department of Biomedical Engineering, Technion - Israel Institute of Technology, Haifa, Israel.}
\address{$^2$ Division of Neurology, The Hospital for Sick Children and Department of Paediatrics, University of Toronto, Toronto, Canada.}
\address{$^3$ Department of Electrical Engineering (ESAT), Stadius Division, KU Leuven, Leuven, Belgium.}
\address{$^4$ Department of Pediatrics, Section of Neurology, Alberta Children’s Hospital and University of Calgary, Calgary, Canada.}
\address{$^5$ Applied Health Research Centre (AHRC), Li Ka Shing Knowledge Institute of St. Michael’s Hospital, Canada.}
\address{$^6$ Ruth and Bruce Rappaport Faculty of Medicine, Technion – Israel Institute of Technology, Haifa, Israel.}
\address{*These authors share senior authorship.}
\ead{jbehar@bm.technion.ac.il}
\vspace{10pt}
\begin{indented}
\item[]April 2022
\end{indented}

\begin{abstract}
\\
\textbf{Objective:} 
Epileptic seizures are relatively common in critically-ill children admitted to the pediatric intensive care unit (PICU) and thus serve as an important target for identification and treatment. Most of these seizures have no discernible clinical manifestation but still have a significant impact on morbidity and mortality. Children that are deemed at risk for seizures within the PICU are monitored using continuous-electroencephalogram (cEEG). cEEG monitoring cost is considerable and as the number of available machines is always limited, clinicians need to resort to triaging patients according to perceived risk in order to allocate resources. This research aims to develop a computer aided tool to improve seizures risk assessment in critically-ill children, using an ubiquitously recorded signal in the PICU, namely the electrocardiogram (ECG).
\\
\textbf{Approach:} 
A novel data-driven model was developed at a patient-level approach, based on features extracted from the first hour of ECG recording and the clinical data of the patient.
\\
\textbf{Main results:}
The most predictive features were the age of the patient, the brain injury as coma etiology and the QRS area. For patients without any prior clinical data, using one hour of ECG recording, the classification performance of the random forest classifier reached an area under the receiver operating characteristic curve (AUROC) score of 0.84. When combining ECG features with the patients clinical history, the AUROC reached 0.87.
\\
\textbf{Significance:}
Taking a real clinical scenario, we estimated that our clinical decision support triage tool can improve the positive predictive value by more than 59\% over the clinical standard.
\end{abstract}

\vspace{0.5pc}

\ioptwocol

\section{Introduction}
\label{sec:introduction}

A seizure is defined as a transient occurrence of signs and/or symptoms due to abnormal excessive or synchronous neuronal activity in the brain \cite{fisher2005epileptic}. In the pediatric intensive care unit (PICU), the prevalence of epileptic seizures among all children admitted is estimated to be 0.8 \% \cite{valencia2006epileptic}. The prevalence among children presenting with reduced level of consciousness or acute brain injury is much higher. Most of these seizures have no discernible clinical correlate (motor movement). These seizures are termed non-convulsive or subclinical seizures \cite{payne2014seizure} and are traditionally detected using continuous-electroencephalogram (cEEG) monitoring that is used on suspected children.

\begin{figure}[!b]
\centerline{\includegraphics[width=\columnwidth]{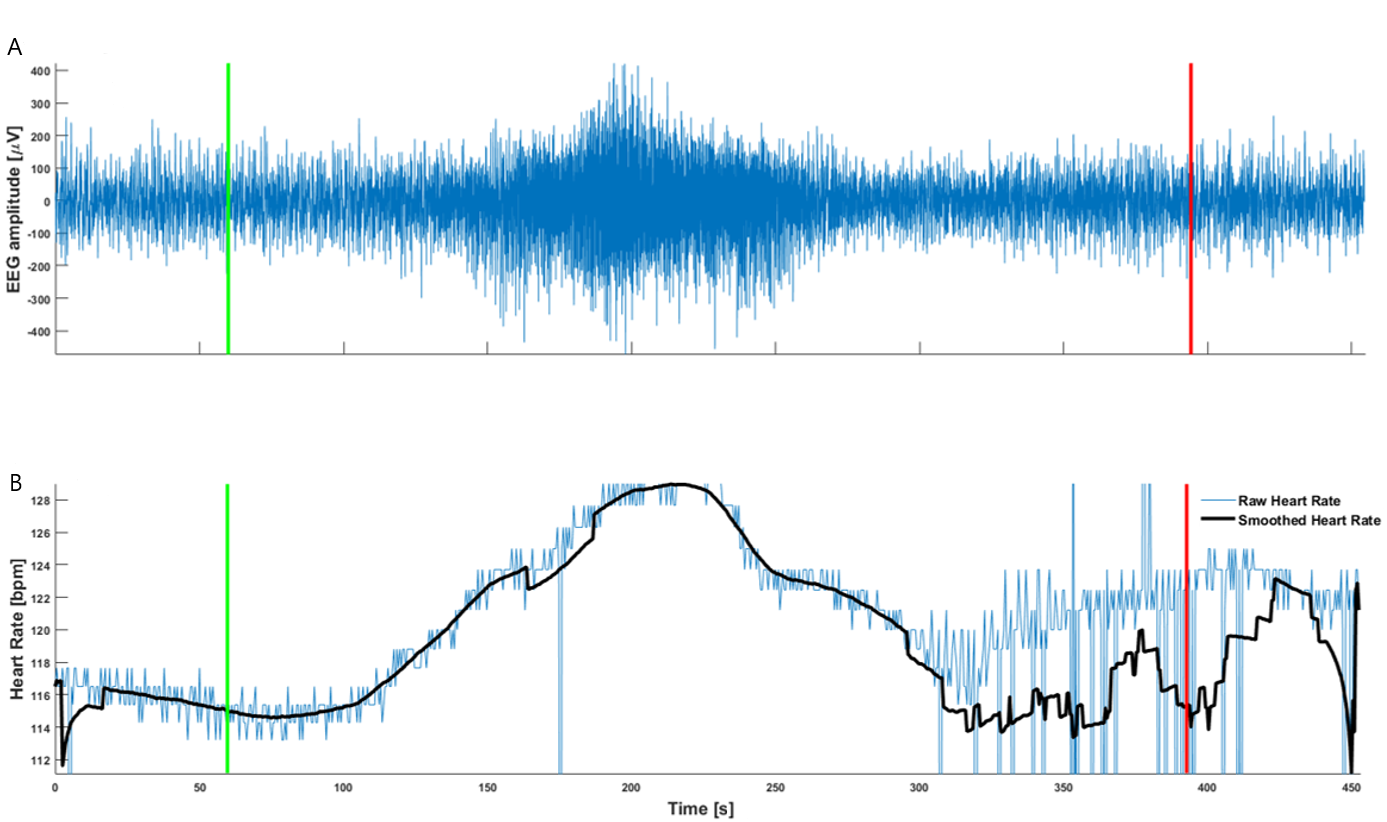}}
\caption{\footnotesize{EEG and heart rate during seizure event. Panel A: EEG recording. Panel B: Heart rate. Green line: start of the seizure. Red line: end of the seizure.}}
\label{fig:EEG_ECG}
\end{figure}

\begin{figure}[!t]
\centerline{\includegraphics[width=\columnwidth]{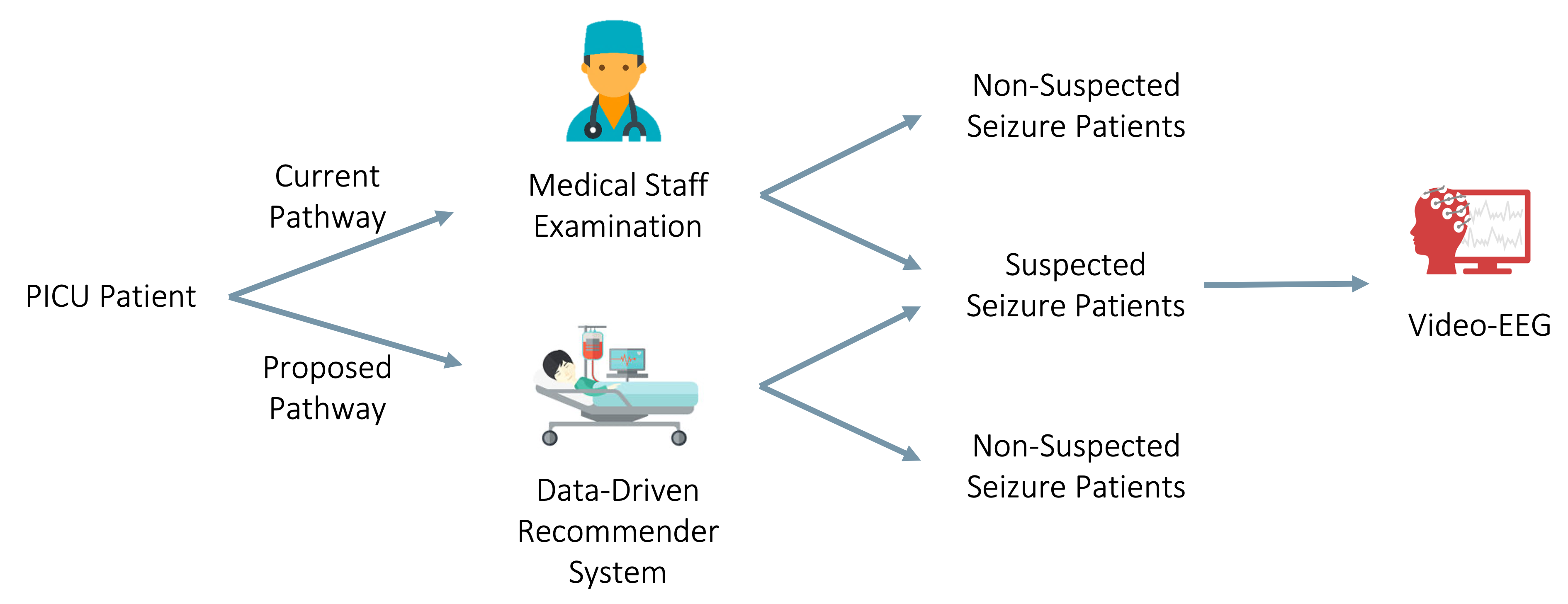}}
\caption{\footnotesize{Current clinical pathway and newly suggested pathway using the data-driven approach for triage of children with epileptic seizures in the pediatric intensive care unit (PICU).}}
\label{fig:Pathway}
\end{figure}

The detection of these subclinical seizures is crucial since these seizures have a significant impact on morbidity, mortality and neuronal health \cite{connell,legido}. The treatment of these seizures using anti-epileptic medication reduces the duration of seizure patterns resulting into a possible reduction of brain injury as observed in \cite{van2010effect}. The use of cEEG has grown in the PICU and became a standard of care in many North American healthcare centers \cite{sanchez2013pediatric}. 
Among the children suspected of epileptic seizures and monitored using cEEG, status epilepticus, defined as a seizure that lasts more than 5 minutes \cite{lowenstein1999s}, is established for 9-32\% \cite{vakorin2021alterations} of the monitored patients. In addition, other patients that are not suspected and not monitored with cEEG might be missed. Given the considerable cost of the cEEG equipment as well as the staff cost for setting up and reviewing the cEEG traces, there is a high interest in developing better triage methods that identifies children at risk of presenting epileptic seizures and that would benefit from continuous cEEG monitoring. This would enable better resource allocation.

According to literature \cite{nei2009cardiac} and as can be observed in Fig. \ref{fig:EEG_ECG}, there is a physiological cardiac effect of the seizure event on the heart rate. This is the reason why numerous studies attempted to perform seizure detection using the ECG of the patient, notably the work of Osorio et al. \cite{osorio2014automated} that focused on finding an increase in the heart rate with the onset of epileptic seizure. The false positive ratio reached by the algorithm of Osorio was between 1.1/h and 9.5/h for a sensitivity of 86\% and 98\% respectively. In addition to the high number of false positives, only seizures with behavioral changes (clinical seizures) were analyzed. In the present research, the majority of seizures in the database are subclinical (Fig. \ref{fig:Manifestation}).

We hypothesized that children experiencing epileptic seizures will present specific patterns in their ECG recordings, either in the signal's morphology or in the beat-to-beat interval statistics. ECG recordings are simple, noninvasive and are routinely and continuously recorded in all children in the PICU. Thus, if seizure specific patterns are found in the ECG it will be possible to mass triage children presenting seizures in the PICU and thus improve their clinical management (Fig. \ref{fig:Pathway}).
\section{Methods}
\label{sec:methods}
\subsection{Database Description}
\label{sec:database}
\begin{figure}[!t]
\centerline{\includegraphics[width=\columnwidth]{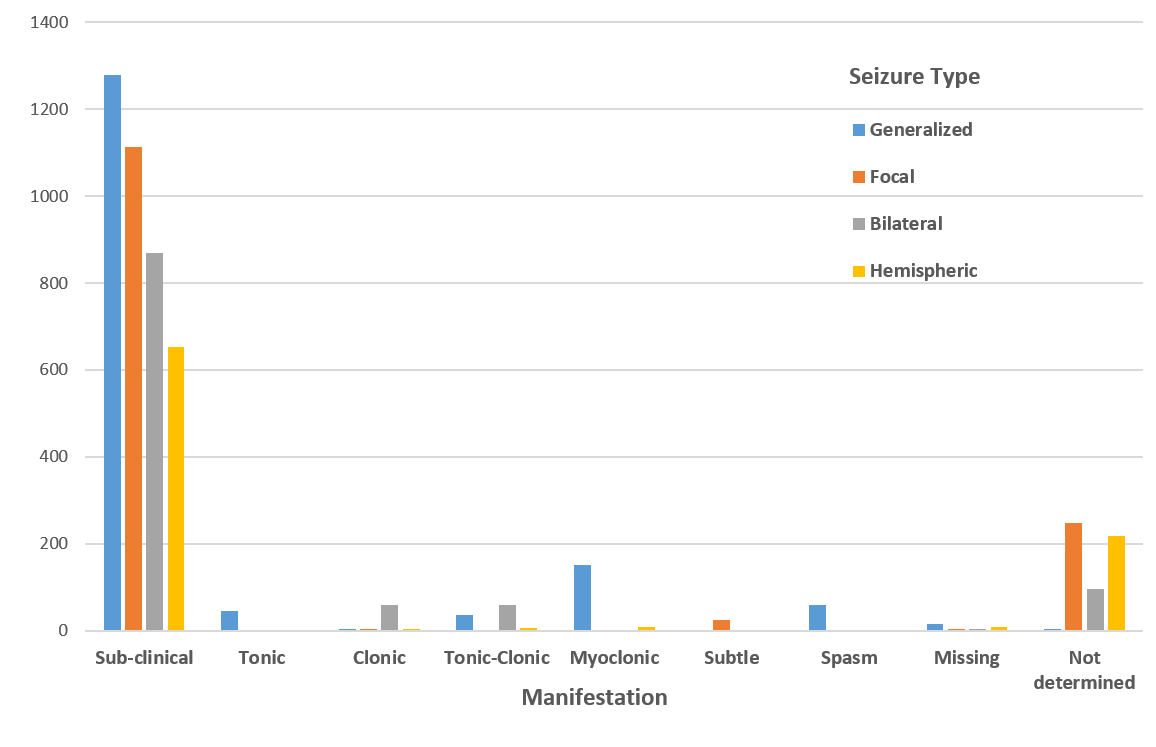}}
\caption{\footnotesize{Seizure types and clinical manifestations.}}
\label{fig:Manifestation}
\end{figure}

The database includes synchronous cEEG and ECG recordings of 176 children with a reduced level of consciousness. More details regarding this database were presented in a previous study \cite{vakorin2021alterations} and data usage was granted  under SickKids Research Ethics Board number 1000014179. Among these patients, 52 children experienced seizures of any type and any length and 124 children were without any seizures. The interquartile range of the age of the patients were between 0.8 and 11.5 years with a median age of 5.1 years. The gender distribution of the study cohort was of 47\% male and 53\% female. All children were admitted between 2010 and 2013 to the PICU at The Hospital for Sick Children in Toronto, Canada and were continuously monitored using cEEG and ECG. Seizures annotations were obtained by reviewing the cEEG and delineation of the beginning and end of the seizures. The annotations were performed by a single board-certified pediatric neurologist. A total of 4,969 reference seizure annotations were obtained. The ECG channel was extracted from the full database that includes recordings of the 176 patients divided in 1182 parts for a total of 10,921 hours of recording. Exploration revealed that the database is highly-imbalanced with respect to the manifestation of the seizures with an over-representation of subclinical seizures as can be seen in Fig. \ref{fig:Manifestation}. Indeed, 79\% of all the seizures in this database are subclinical. This is an important point since the detection of these seizures is more challenging due to an absence of clinical manifestation.

\subsection{Preprocessing}
The first step of the process is R-peak detection. The ECG signal was filtered using a band-pass zero phase 5th order Butterworth filter with a passband from 3 to 45 Hz. R-peak detection was performed using jqrs \cite{behar2014comparison} with a threshold at 0.5 (n.u.) and a refractory period of 150msec. This is smaller than the traditional 250ms refractory period used in adult R-peak detection because of children's higher heart rate. The detected R-peak were then adjusted by looking for the local minimum within a 50ms window around the detected fiducial. For each patient and segment the NN interval was computed using filtrr \cite{behar2018physiozoo} in order to filter out suspected ectopic beats, missed beats and artifacts with a combination of range, moving average and quotient filter.

\subsection{Local Approach}
\begin{figure}[!t]
\centerline{\includegraphics[width=\columnwidth]{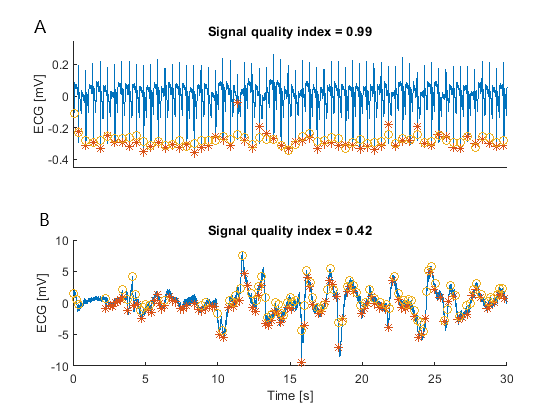}}
\caption{\footnotesize{Example of R-peak detection on the ECG recording signal using two R-peak detectors to estimate signal quality. Panel A: ECG with high signal quality index. Panel B: ECG with low signal quality index.}}
\label{fig:bSQI}
\end{figure}

The aim of this approach is to detect individual events using the state-of-the-art algorithm presented in the introduction (\ref{sec:introduction}). The algorithm used for seizure detection is the Osorio algorithm \cite{osorio2014automated}. For each recording the Osorio algorithm was used on the RR interval extracted from the ECG with two sets of parameters. For the first set of parameters, a seizure is detected if the heart rate increases in 15\% compared to the baseline. For the second set, the heart rate has to increase by 30\% for at least 5 seconds. Each detected event is associated with a reference event and then considered as true positive if it occurs within a window of 60 seconds before the beginning and 60 seconds after the end of the seizure. Finally, in order to reduce the number of false alarms due to signal quality, the detection that occurs in region of low quality signal were discarded. To perform the signal quality estimation a second R-peak detector from PhysioZoo HRV called rqrs is used \cite{behar2018physiozoo}. The two sequences of R-peak were then compared using the bSQI quality index \cite{behar2013ecg}. In the case that the bSQI at the time of the detected event within a window of 60 seconds is lower than 0.8, the detection is discarded (Fig. \ref{fig:bSQI}).

\subsection{Global Approach}
For this second approach, the machine learning problem is framed so to classify a patient as ``Seizure Patient'' or ``Non Seizure Patient'' versus detecting specific events as in the ``Local approach''. Thus, the classification is done at the patient level and the assumption is that the model will be less vulnerable to repeated false alarms. The ``Seizure Patient'' label is given to patients that have a seizure that lasts more than 5 minutes within the first 48 hours of recording. This is inspired by the medical definition  of Status Epilepticus \cite{lowenstein1999s}, indicating a clinically-relevant threshold of seizure burden. In order to evaluate the clinical usability of this system within a hospital setting the triage system must be able to flag patients with risk for seizures shortly after their admission. As noted above, we hypothesized that a seizure patients’ ECG will contain discernible information even outside of active seizure periods. Thus, we used only the first hour of recording for inference on the test set, simulating the clinical scenario of using this tool to triage patients according to seizure risk in order to guide medical decision support regarding cEEG monitoring.

\begin{figure}[!b]
\centerline{\includegraphics[width=\columnwidth]{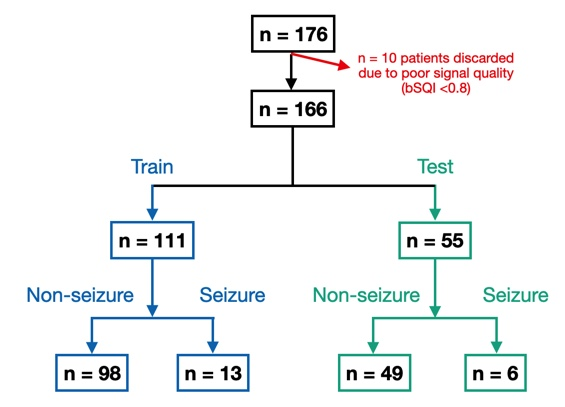}}
\caption{\footnotesize{Division of the patients (n) into train and test set with stratification after signal quality check.}}
\label{fig:Split}
\end{figure}

\subsection{Features Engineering}
Three different categories of features were considered: the clinical information of the patient, also called ``metadata'' (META), the Heart Rate Variability (HRV) features \cite{behar2018physiozoo} and the ECG morphological features (MOR) \cite{biton2021atrial}. The 16 META features include clinical data about the patient such as standard demographics (i.e. age, gender) and other clinical variables that can be used to triage the patients at risk of seizure (e.g. prior diagnosis, presumed cause of coma). The full list of META features appears in Table \ref{tab:META}. The second category of features are features based on the HRV. These features were extracted using the PhysioZoo HRV toolbox leading to 25 HRV features (Table \ref{tab:HRV}). In order to extract ECG morphological features, the first step is to detect the different parts of the ECG signal. The detection of the fiducial points on the ECG waveform was done using the popular open source wavedet algorithm \cite{martinez2004wavelet}. A total of 74 features \cite{biton2021atrial} were extracted from these fiducial points: 38 features extracted from interval duration (Table \ref{tab:MOR1}) and 36 from waves characteristics (Table \ref{tab:MOR2}). 

\begin{table}[]
\centering
\caption{\footnotesize{List of META features.}}
\begin{tabular}{p{45pt}|p{170pt}}\hline
 \textbf{Name} & \textbf{Definition}                                                       \\ \hline
 Age           & Age                                                \\
Gender        & Gender                                                                 \\
 DevDelay      & Developmental delay prior to PICU admission.                         \\
 Epilepsy      & Epilepsy prior to PICU admission.                                    \\
 PreDxNa       & Any patient prior diagnoses.                                         \\
 Seizure       & Clinical Seizures as part of their acute presentation.               \\
 EtSeiz        & Are seizures a presumed coma etiology.                                \\
 SeizPrim      & Are seizures the principal cause of coma .                     \\
 EtBrainInj    & Is brain injury a presumed coma etiology.           \\
 BrainInjPrim  & Is brain injury the principal cause of coma. \\
 EtMetab       & Is metabolic or toxic illness a presumed coma etiology.                      \\
 MetPrim       & Is metabolic or toxic illness the principal cause of coma.            \\
 EtSyst        & Is systemic illness a presumed coma etiology.                        \\
 SysPrim       & Is systemic illness the principal cause of coma.              \\
 EtSed         & Are sedating medications a presumed coma etiology.                   \\
 SedPrim       & Are sedating medications the principal cause of coma.         \\ \hline
\end{tabular}
\label{tab:META}
\end{table}

\begin{table}[!t]
\centering
\caption{\footnotesize{List of HRV features.}}
\begin{tabular}{p{40pt}|p{175pt}}\hline
 \textbf{Name}          & \textbf{Definition}  \\\hline

 bSQI          & Signal quality index [n.u.]   \\
 Alpha 1       & DFA low-scale slope [n.u.]   \\
 Alpha 2       & DFA high-scale slope [n.u.]   \\
 AVNN          & Average NN interval duration [ms]     \\
 BETA          & Slope of the linear interpolation of the spectrum in a log-log scale for frequencies below the upper bound of the VLF band. [n.u.] \\  
 HF norm       & High frequency power in normalized units [\%] \\
 HF peak       & Peak frequency in the high frequency band [Hz]     \\
 HF power      & Power in the high frequency band [ms2]\\
 IALS          & Inverse average length of the acceleration/deceleration segments [n.u.] \\
 LF norm       & Low frequency power in normalized units [\%] \\
 LF peak       & Peak frequency in the low frequency band [Hz] \\
 LF power      & Power in the low frequency band [ms2]    \\
 LF to HF      & Low frequency band to high frequency band power ratio [n.u.]   \\
 PAS           & The percentage of NN intervals in alternation segments [\%]     \\
 PIP           & Percentage of inflection points in the NN interval time series [\%]     \\
 pNN50         & Percent of NN interval differences greater than 50 milliseconds [\%]     \\
 PSS           & Percentage of short segments [\%]     \\
 RMSSD         & The square root of the mean of the sum of the squares of differences between adjacent NN intervals [ms]     \\
 SampEn        & Sample entropy [n.u.]   \\
 SD1           & NN interval standard deviation along the perpendicular to the line-of-identity [ms]     \\
 SD2           & NN interval standard deviation along the   line-of-identity [ms]     \\
 SDNN          & Standard deviation of NN interval duration [ms]     \\
 SEM           & Standard error of the mean NN interval [ms]     \\
 Total power   & Total power [ms2]    \\
 VLF norm      & Very low frequency power in normalized units [\%]     \\
 VLF power     & Power in the very low frequency band [ms2] \\\hline  
\end{tabular}
\label{tab:HRV}
\end{table}

\subsection{Machine Learning}
\subsubsection{Data preparation}
All one-hour recordings with a mean bSQI lower than 0.8 were discarded and only recordings with a high enough quality score were analyzed. Due to a bSQI lower than 0.8 for all the signals in the first 48 hours of recording of 10 patients, the entire recording of these patients was discarded. The 166 remaining patients were divided into train and test set with a ratio of 2/3 (111 patients) and 1/3 (55 patients) respectively as represented in Fig. \ref{fig:Split}. The division was done with stratification resulting in the same ratio of seizure and non-seizure patients in the train and test set.
HRV and MOR features were computed on the whole hour of data for each hour of recording. In order to augment the training database, the 48 first hours of recording were used resulting into 3,667 examples in the training set (3144 non-seizure examples and 523 seizure examples). However, for the test set only the first hour of the recordings were considered for inference to simulate a clinical scenario. Indeed, in order to be clinically applicable the data-driven model must be able to provide a recommendation on triage soon after the child is hospitalized. The features were normalized using a Min-Max scaler that transform each feature by scaling them between zero and one. This scaler is fitted on the training set and applied in both the training and test set.

\subsubsection{Training strategy}
A random forest (RF) classifier was trained. Since the number of patients in the dataset is limited, one can check the algorithm’s performance across different train-test splits in order to obtain more significant statistics performance, a method called nested cross-validation. The idea behind this method is to evaluate the generalization error within two different loops. The inner loop, in which the train set is divided into a train and validation set, is used to the hyper-parameters optimization and the outer loop for assessing the model performance over different test sets. In this work, the train-test split was performed 10 times and at each time the division was done through random sub-sampling and with stratification. 

\subsubsection{Feature Selection}
We used the recursive feature elimination algorithm \cite{guyon2002gene} for feature selection. This algorithm performs features selection by recursively pruning the features with the lowest feature importance. The model is first trained with all the available features and, at each iteration, one feature is eliminated until the desired number of features is reached.  We set the number of features to be selected at each train-test split to 9 corresponding to approximately 10\% of the number of patients in the train set according to the well-known ``rule of thumb'' that states that one should have at least 10 examples for one feature. 

\subsubsection{Bayesian Search of Hyperparameter tuning}
A total of 300 estimators were used. The number of estimators was not considered as an hyper-parameter since increasing the number of estimators to use in a RF Classifier reduces the model error at the cost of higher training time \cite{probst2017tune}. Other important hyper-parameters were selected using a Bayesian search strategy \cite{wu2019hyperparameter} and a 5-fold stratified cross-validation with an Area Under the Receiver Operating Characteristic curve (AUROC) scoring metric. The search range is summarized in Table \ref{tab:hyper}. Concerning the other parameters of the RF classifier, the default values were chosen: min\_samples\_split=2, min\_samples\_leaf=1, max\_leaf\_nodes=None, max\_sample=None.

\subsubsection{Machine Learning Models}
In order to highlight the contribution of each type of feature, four different models were considered. The first one included the age of the patient only. The second one, denoted META, included the 16 clinical features available (Table \ref{tab:META}). The third one had 101 features and included the age of the patient and all the features extracted from the ECG recording i.e. HRV features \cite{behar2018physiozoo} and MOR features \cite{biton2021atrial}. The last model included all the 116 available features (META, HRV and MOR). The same train-test splits, feature selection procedure and hyperparameters optimization were performed for all four models.

\subsection{Performances Measures}
Keeping in mind the clinical used-case scenario of a decision-support tool, the two performance measures that were used to assess the performance of the local approach, i.e. detection of individual seizure events, were the Sensitivity (Se) and the Positive Predictive Value (PPV), given by:

\begin{equation} Se = \frac{TP}{TP+FN} \label{eq1}\end{equation}
\begin{equation} PPV = \frac{TP}{TP+FP} \label{eq2}\end{equation}

Where TP stands for True Positive, FN for False Negative and FP for False Positive.
For the global approach, i.e. detection of seizure patients, in addition to Se and PPV, the classification performance was quantified using the AUROC score which reflects the level of separability between the two classes (i.e. seizure patient, non-seizure patient). 

\section{Results}
\label{sec:results}
\subsection{Local Approach}

\begin{table}[!t]
\centering
\caption{\footnotesize{Performance statistics of the Osorio algorithm \cite{osorio2014automated}. Se and PPV are in percent.}}
\begin{tabular}{c|ccccc}
\hline
                  & \textbf{TP} & \textbf{FP} & \textbf{FN} & \textbf{Se} & \textbf{PPV} \\ \hline
\textbf{Osorio15} & 549         & 21438       & 3577        & 13                  & 2.5                  \\
\textbf{Osorio30} & 52          & 1041        & 4074        & 1.2                 & 4.8                  \\ \hline
\end{tabular}
\label{tab:local} 
\end{table}

\begin{table}[!t]
\centering
\caption{\footnotesize{AUROC performance obtained for the four models.}}
\begin{tabular}{c|cc|cc}
\hline
\multicolumn{1}{c|}{\textbf{Model}} & \multicolumn{2}{c|}{\textbf{AUROC train}}     & \multicolumn{2}{c}{\textbf{AUROC test}}       \\ \hline
\multicolumn{1}{l|}{}      & Mean                  & Std                   & Mean                  & Std                   \\
Age                        & 0.95                  & 0.02                  & 0.81                  & 0.05                  \\
META                       & 0.95                  & 0.02                  & 0.85                  & 0.02                  \\
Age+HRV+MOR                    & 0.94 & 0.02 & 0.84 & 0.05 \\

META+HRV+MOR                   & 0.96 & 0.02 & 0.87 & 0.04
\\ \hline
\end{tabular}
\label{tab:roc}
\end{table}

Table \ref{tab:local} presents the performance statistics of the Osorio \cite{osorio2014automated} algorithm with the different set of parameters. Osorio15 and Osorio30 refer to use of the Osorio algorithm with the first and second set of parameters, i.e. an increase of 15\% and an increase of 30\% during 5 seconds, accordingly. Referring to Table \ref{tab:local}, the first set of parameters leads to a higher number of detected seizures (i.e. TP) and lower number of missed seizures (i.e. FN) but also to a higher number of FP than the second set of parameters. Since the first set of parameters is more sensitive, the Se reached (13\%) is higher than for the second set of parameters (1.2\%). However, the PPV obtained was higher with the less sensitive parameters. Despite that, the PPV in this case was approximately 5\%, that is to say that only 1 of 20 detected events was actually a real seizure. 

\begin{figure}[!b]
\centerline{\includegraphics[width=\columnwidth]{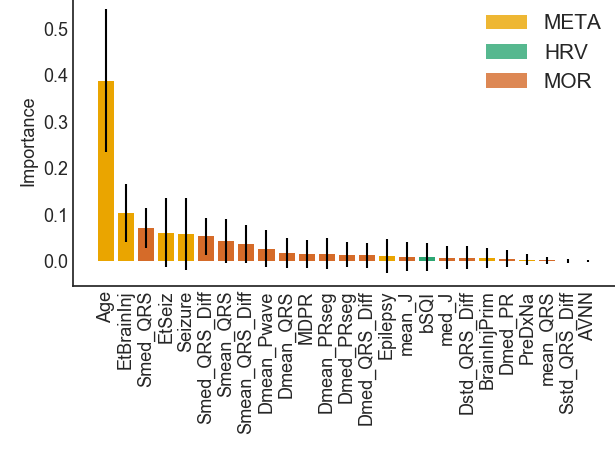}}
\caption{\footnotesize{Mean and standard deviation of the features importance over the different train-test split of the META+HRV+MOR model.}}
\label{fig:Importance_All}
\end{figure}

\subsection{Global Approach}
Table \ref{tab:roc} presents the mean and standard deviation of the AUROC score over the 10 different train-test splits obtained for the four models. As one can see, the META+HRV+MOR model obtained the best score with a mean AUROC of 0.87 on the test set. The age had a sensible importance in the classification of the patients as can be seen through features importance of the META+HRV+MOR model (Fig. \ref{fig:Importance_All}). However, the use of the age only led to notably lower performances with a mean AUROC of 0.81 on the test set. Concerning the Age+HRV+MOR model, the AUROC reached was 0.84, highlighting the contribution of the ECG based features to this classification task. The same observation could be done on the META (AUROC of 0.85) and META+HRV+MOR (AUROC of 0.87)models.

\subsection{Usage of the Developed Algorithms and Clinical Scenario}
The first interesting model that can be used is the META+HRV+MOR model that reach the highest AUROC. In this case, the data-driven recommender system mentioned in Fig. \ref{fig:Pathway} is based on the 16 clinical data features, in addition to the HRV and MOR features extracted from the first hour of recording. However, this model is not usable without the clinical data of the patient. In many cases these may not be available (e.g. first hospitalization of the patient). For this reason, a model working from the raw ECG is of interest. The Age+HRV+MOR model is usable after one hour of ECG recording and therefore represents a viable solution that does not need any medical staff participation except for ECG placement which is ubiquitous as part of routine PICU care. 

\begin{figure}[!b]
\centerline{\includegraphics[width=\columnwidth]{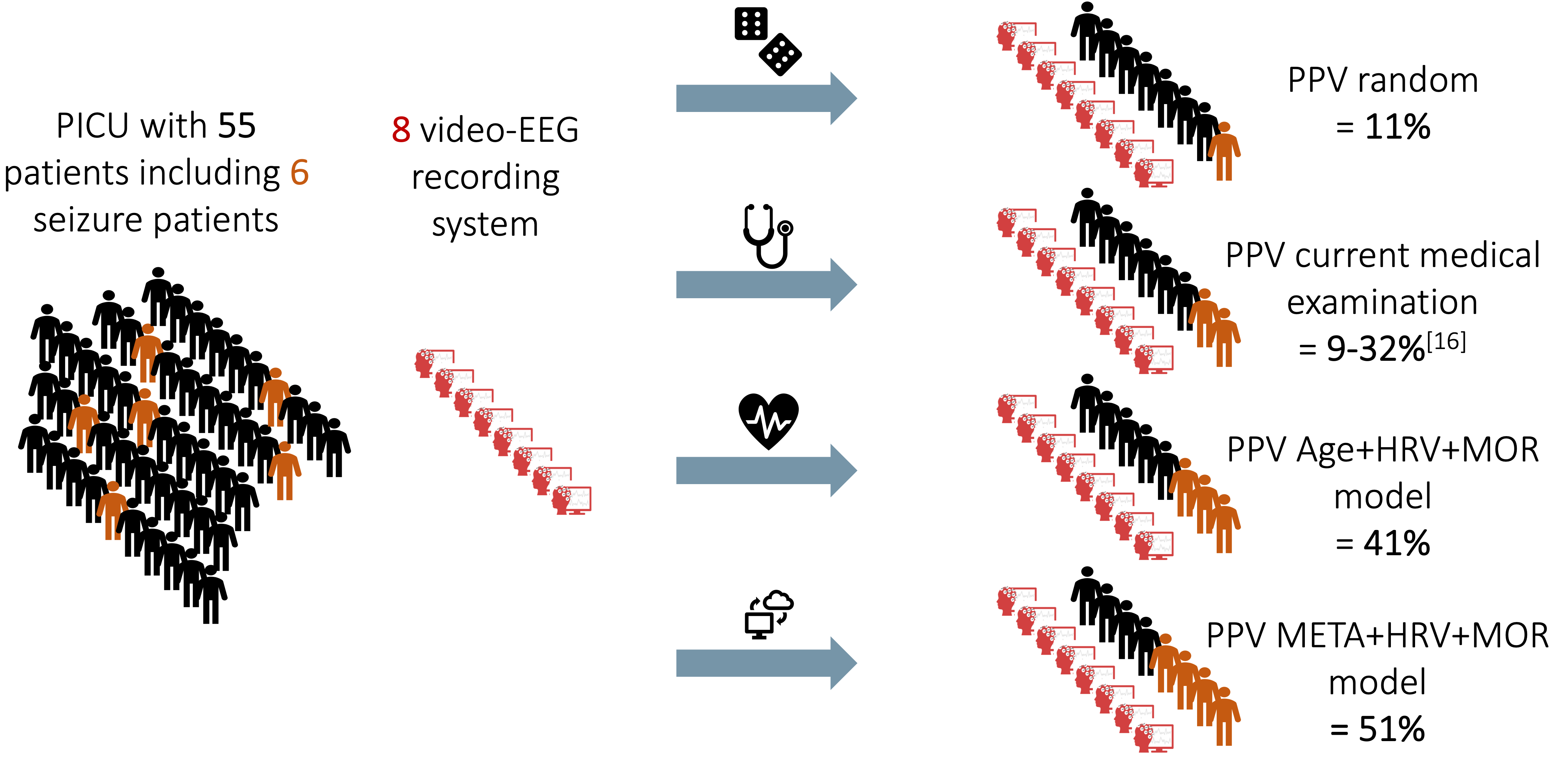}}
\caption{\footnotesize{Comparison between the PPV reached in a clinical scenario by: a random classification, the current medical practice, the Age+HRV+MOR model, and the META+HRV+MOR model.}}
\label{fig:Clinical_scenario}
\end{figure}

In order to evaluate the classification performances of the developed algorithms, one can consider the scenario of a PICU with 55 patients including 6 seizure patients, similar  to the test-set of this work (Fig. \ref{fig:Split}), and we assume that 8 cEEG systems are available. In this setting, a random distribution of the 8 cEEG systems over the patients leads to a PPV of 11\%. In the current medical examination pathway, among the children suspected of epileptic seizures and monitored using cEEG, only 9-32\% of them are reported as presenting epileptic seizure events \cite{vakorin2021alterations}. Thus, the best reported PPV of the clinical risk stratification system is 32\%. To compute the PPV reached by the data-driven algorithms, one first need to consider a threshold that can be used to transform the prediction probabilities of each example into a binary prediction. This threshold was established according to the clinical setting. Since 8 cEEG systems were available, the 8 patients with the highest prediction probabilities were considered as ``positive prediction''. The resulting PPV of the Age+HRV+MOR model was 41\% and the PPV of the META+HRV+MOR model was 51\%. In other words, the Age+HRV+MOR and the META+HRV+MOR models achieved a relative improvement of 28\% and 59\% respectively compared to the best clinical practice PPV as reported in \cite{vakorin2021alterations}. Fig. \ref{fig:Clinical_scenario} illustrates the usage of the developed algorithms in a clinical scenario.

\section{Discussion}
\label{sec:discussion}
The local approach led to a high number of false positives and a low sensitivity. From a clinical point of view, this high number of false alarms make this approach impractical as a triage tool, and thus, our results support that a global approach, i.e. classification at the patient level as seizure or non-seizure, may be more appropriate and clinically useful. Indeed, the global approach obtained a better performance with AUROC=0.87 for the best model (META+HRV+MOR) and AUROC=0.84 for the Age+HRV+MOR model. This last model is particularly interesting because it has no need for clinical information about the patient that often is hard to acquire early into the patient's stay in the critical care.

\begin{figure}[!b]
\centerline{\includegraphics[width=\columnwidth]{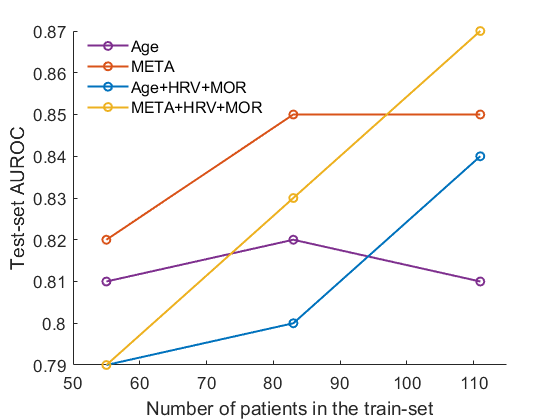}}
\caption{\footnotesize{Learning curves. Mean test-set AUROC in function of patients in the training set.}}
\label{fig:Learning_curves}
\end{figure}

The age was highly associated with the appearance of seizures (Fig. \ref{fig:Importance_All}). As one can see in Fig. \ref{fig:Age_hist} younger patients were more susceptible to epileptic seizures. This result is in accordance with previously published studies \cite{fogarasi2002effect} that hypothesize that events in brain maturation in childhood significantly affect the propensity to suffer epileptic seizures. Another important clinical feature was the presumption of coma etiology caused by brain injury. This relation between brain injury and the prevalence of epileptic seizures was observed in many studies \cite{mccoy2011predictors,abend2013electrographic}. Concerning the ECG based features, Fig. \ref{fig:Importance_All} revealed that features related to the QRS complex interval area were found to be significant in this classification task. This finding suggests that the QRS complex is altered for seizure patients. Of note, a similar observation was done on patients suffering from arrhythmia \cite{krasteva2006automatic}. A possible confounding factor in the relation between QRS morphology and seizure risk is the patient's age that affects both. As this dataset is too small, we plan to examine this intriguing result in future studies, using a larger cohort.

The learning curves in Fig. \ref{fig:Learning_curves} shows that increasing the size of the training set continuously improves the models performance for models that include features extracted from the ECG signal i.e. Age+HRV+MOR and META+HRV+MOR thus suggesting that the classification performance for these models can be improved further. This is different for the Age and META AUROC scores which plateaued with the size increase of the training set.

\subsection{Error Analysis}
\begin{figure}[h]
\centerline{\includegraphics[width=\columnwidth]{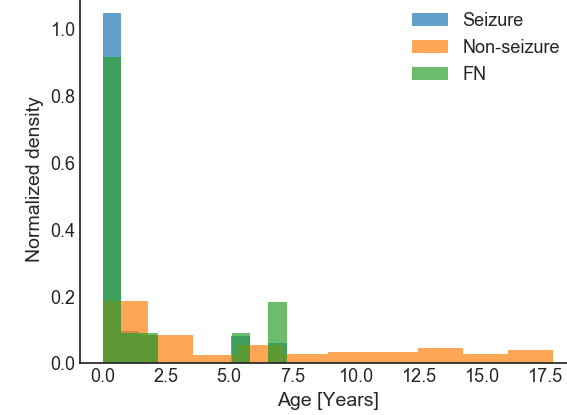}}
\caption{\footnotesize{Histogram of the age of seizure patients (blue), non-seizure patients (orange) and misclassified false negatives (green).}}
\label{fig:Age_hist}
\end{figure}

An interesting observation in Fig. \ref{fig:Age_hist} is that a part of the misclassified false negatives (i.e. missed seizure patients) have a higher age compared to the majority of seizure patients.

\subsection{Limitations}
This work has some limitations. First, even though our results are based on a relatively large dataset compared to previously published studies interested in the task of seizure detection from ECG, it was collected in a single center. Thus, in order to evaluate the generalization performance of the presented algorithms, external test set from other hospitals needs to be collected and included. A second important point is the moderated size of the available dataset. Indeed, as the learning curves showed (Figure \ref{fig:Learning_curves}), the performance of the ECG based models continue improving as the training set grows. Moreover, with a very large database, a deep-learning approach could be considered, possibly increasing further the classification performance. Finally, from a clinical point of view, more physiological time series continuously measured in the PICU may be exploited such as the respiratory signal or the photoplethysmography (PPG). The use of these additional channels might improve performance and will be tested in future work. 
\section{Conclusion}
\label{sec:conclusion}
This is the first research attempting to develop a data-driven, clinical decision-support tool aimed at triaging critically-ill children at risk for epileptic seizures. Our tool is based on a commonly acquired signal in the intensive care unit, the ECG. Furthermore, these models were designed throughout the development process with a clinical outlook at their intended use. Future work includes growing the retrospective database to boost models performance as well as the prospective evaluation of the best model in clinical practice.
\section{Acknowledgments}
The authors wish to acknowledge the contributions of the following additional DETECT study investigators: Helena Frndova, Cristina Go, Nicholas S. Abend, William B. Gallentine, Kendall B. Nash, James S. Hutchison, Christopher S. Parshuram, O. Carter Snead. CDH received funding for this work from the SickKids Foundation, the Canadian Institutes of Health Research and the Physicians Services Incorporated Foundation. SVH and TDC received funding from the Flemish Government (AI Research Program) and are affiliated to Leuven.AI - KU Leuven institute for AI, B-3000, Leuven, Belgium.

\section*{References}

\bibliographystyle{IEEEtran}
\bibliography{references}

\begin{thebibliography}{10}
\providecommand{\url}[1]{#1}
\csname url@samestyle\endcsname
\providecommand{\newblock}{\relax}
\providecommand{\bibinfo}[2]{#2}
\providecommand{\BIBentrySTDinterwordspacing}{\spaceskip=0pt\relax}
\providecommand{\BIBentryALTinterwordstretchfactor}{4}
\providecommand{\BIBentryALTinterwordspacing}{\spaceskip=\fontdimen2\font plus
\BIBentryALTinterwordstretchfactor\fontdimen3\font minus
  \fontdimen4\font\relax}
\providecommand{\BIBforeignlanguage}[2]{{%
\expandafter\ifx\csname l@#1\endcsname\relax
\typeout{** WARNING: IEEEtran.bst: No hyphenation pattern has been}%
\typeout{** loaded for the language `#1'. Using the pattern for}%
\typeout{** the default language instead.}%
\else
\language=\csname l@#1\endcsname
\fi
#2}}
\providecommand{\BIBdecl}{\relax}
\BIBdecl

\bibitem{fisher2005epileptic}
R.~S. Fisher, W.~V.~E. Boas, W.~Blume, C.~Elger, P.~Genton, P.~Lee, and
  J.~Engel~Jr, ``Epileptic seizures and epilepsy: definitions proposed by the
  international league against epilepsy (ilae) and the international bureau for
  epilepsy (ibe),'' \emph{Epilepsia}, vol.~46, no.~4, pp. 470--472, 2005.

\bibitem{valencia2006epileptic}
I.~Valencia, G.~Lozano, S.~V. Kothare, J.~J. Melvin, D.~S. Khurana, H.~H.
  Hardison, S.~S. Yum, and A.~Legido, ``Epileptic seizures in the pediatric
  intensive care unit setting,'' \emph{Epileptic disorders}, vol.~8, no.~4, pp.
  277--284, 2006.

\bibitem{payne2014seizure}
E.~T. Payne, X.~Y. Zhao, H.~Frndova, K.~McBain, R.~Sharma, J.~S. Hutchison, and
  C.~D. Hahn, ``Seizure burden is independently associated with short term
  outcome in critically ill children,'' \emph{Brain}, vol. 137, no.~5, pp.
  1429--1438, 2014.

\bibitem{connell}
J.~Connell, R.~Oozeer, L.~De~Vries, L.~Dubowitz, and V.~Dubowitz, ``Continuous
  eeg monitoring of neonatal seizures: diagnostic and prognostic
  considerations.'' \emph{Archives of disease in childhood}, vol.~64, no. 4
  Spec No, pp. 452--458, 1989.

\bibitem{legido}
A.~Legido, R.~R. Clancy, and P.~H. Berman, ``Neurologic outcome after
  electroencephalographically proven neonatal seizures,'' \emph{Pediatrics},
  vol.~88, no.~3, pp. 583--596, 1991.

\bibitem{van2010effect}
L.~G. van Rooij, M.~C. Toet, A.~C. van Huffelen, F.~Groenendaal, W.~Laan,
  A.~Zecic, T.~de~Haan, I.~L. van Straaten, S.~Vrancken, G.~van Wezel
  \emph{et~al.}, ``Effect of treatment of subclinical neonatal seizures
  detected with aeeg: randomized, controlled trial,'' \emph{Pediatrics}, vol.
  125, no.~2, pp. e358--e366, 2010.

\bibitem{sanchez2013pediatric}
S.~M. Sanchez, J.~Carpenter, K.~E. Chapman, D.~J. Dlugos, W.~Gallentine, C.~C.
  Giza, J.~L. Goldstein, C.~D. Hahn, S.~K. Kessler, T.~Loddenkemper
  \emph{et~al.}, ``Pediatric icu eeg monitoring: current resources and practice
  in the united states and canada,'' \emph{Journal of clinical neurophysiology:
  official publication of the American Electroencephalographic Society},
  vol.~30, no.~2, p. 156, 2013.

\bibitem{lowenstein1999s}
D.~H. Lowenstein, T.~Bleck, and R.~L. Macdonald, ``It's time to revise the
  definition of status epilepticus,'' 1999.

\bibitem{vakorin2021alterations}
V.~A. Vakorin, D.~A. Nita, E.~T. Payne, K.~L. McBain, H.~Frndova, C.~Go,
  U.~Ribary, N.~S. Abend, W.~B. Gallentine, K.~B. Nash \emph{et~al.},
  ``Alterations in coordinated eeg activity precede the development of seizures
  in comatose children,'' \emph{Clinical Neurophysiology}, vol. 132, no.~7, pp.
  1505--1514, 2021.

\bibitem{nei2009cardiac}
M.~Nei, ``Cardiac effects of seizures,'' \emph{Epilepsy Currents}, vol.~9,
  no.~4, pp. 91--95, 2009.

\bibitem{osorio2014automated}
I.~Osorio, ``Automated seizure detection using ekg,'' \emph{International
  journal of neural systems}, vol.~24, no.~02, p. 1450001, 2014.

\bibitem{behar2014comparison}
J.~Behar, A.~Johnson, G.~D. Clifford, and J.~Oster, ``A comparison of single
  channel fetal ecg extraction methods,'' \emph{Annals of biomedical
  engineering}, vol.~42, no.~6, pp. 1340--1353, 2014.

\bibitem{behar2018physiozoo}
J.~A. Behar, A.~A. Rosenberg, I.~Weiser-Bitoun, O.~Shemla, A.~Alexandrovich,
  E.~Konyukhov, and Y.~Yaniv, ``Physiozoo: a novel open access platform for
  heart rate variability analysis of mammalian electrocardiographic data,''
  \emph{Frontiers in physiology}, vol.~9, p. 1390, 2018.

\bibitem{behar2013ecg}
J.~Behar, J.~Oster, Q.~Li, and G.~D. Clifford, ``Ecg signal quality during
  arrhythmia and its application to false alarm reduction,'' \emph{IEEE
  transactions on biomedical engineering}, vol.~60, no.~6, pp. 1660--1666,
  2013.

\bibitem{biton2021atrial}
S.~Biton, S.~Gendelman, A.~H. Ribeiro, G.~Miana, C.~Moreira, A.~L.~P. Ribeiro,
  and J.~A. Behar, ``Atrial fibrillation risk prediction from the 12-lead ecg
  using digital biomarkers and deep representation learning,'' \emph{European
  Heart Journal-Digital Health}, 2021.

\bibitem{martinez2004wavelet}
J.~P. Mart{\'\i}nez, R.~Almeida, S.~Olmos, A.~P. Rocha, and P.~Laguna, ``A
  wavelet-based ecg delineator: evaluation on standard databases,'' \emph{IEEE
  Transactions on biomedical engineering}, vol.~51, no.~4, pp. 570--581, 2004.

\bibitem{guyon2002gene}
I.~Guyon, J.~Weston, S.~Barnhill, and V.~Vapnik, ``Gene selection for cancer
  classification using support vector machines,'' \emph{Machine learning},
  vol.~46, no.~1, pp. 389--422, 2002.

\bibitem{probst2017tune}
P.~Probst and A.-L. Boulesteix, ``To tune or not to tune the number of trees in
  random forest.'' \emph{J. Mach. Learn. Res.}, vol.~18, no.~1, pp. 6673--6690,
  2017.

\bibitem{wu2019hyperparameter}
J.~Wu, X.-Y. Chen, H.~Zhang, L.-D. Xiong, H.~Lei, and S.-H. Deng,
  ``Hyperparameter optimization for machine learning models based on bayesian
  optimization,'' \emph{Journal of Electronic Science and Technology}, vol.~17,
  no.~1, pp. 26--40, 2019.

\bibitem{fogarasi2002effect}
A.~Fogarasi, H.~Jokeit, E.~Faveret, J.~Janszky, and I.~Tuxhorn, ``The effect of
  age on seizure semiology in childhood temporal lobe epilepsy,''
  \emph{Epilepsia}, vol.~43, no.~6, pp. 638--643, 2002.

\bibitem{mccoy2011predictors}
B.~McCoy, R.~Sharma, A.~Ochi, C.~Go, H.~Otsubo, J.~S. Hutchison, E.~G. Atenafu,
  and C.~D. Hahn, ``Predictors of nonconvulsive seizures among critically ill
  children,'' \emph{Epilepsia}, vol.~52, no.~11, pp. 1973--1978, 2011.

\bibitem{abend2013electrographic}
N.~S. Abend, D.~H. Arndt, J.~L. Carpenter, K.~E. Chapman, K.~M. Cornett, W.~B.
  Gallentine, C.~C. Giza, J.~L. Goldstein, C.~D. Hahn, J.~T. Lerner
  \emph{et~al.}, ``Electrographic seizures in pediatric icu patients: cohort
  study of risk factors and mortality,'' \emph{Neurology}, vol.~81, no.~4, pp.
  383--391, 2013.

\bibitem{krasteva2006automatic}
V.~T. Krasteva, I.~I. Jekova, and I.~I. Christov, ``Automatic detection of
  premature atrial contractions in the electrocardiogram,''
  \emph{Electrotechniques Electronics E \& E}, vol.~9, no.~10, 2006.

\bibitem{mao2019automated}
L.~Mao, H.~Chen, J.~Bai, J.~Wei, Q.~Li, and R.~Zhang, ``Automated detection of
  first-degree atrioventricular block using ecgs,'' in \emph{International
  Conference on Health Information Science}.\hskip 1em plus 0.5em minus
  0.4em\relax Springer, 2019, pp. 156--167.

\end{thebibliography}
\section*{Supplement}

\begin{table}[h]
\centering
\caption{\footnotesize{Hyper-parameters search range for the RF model.}}
\begin{tabular}{p{130pt}|p{80pt}}\hline
\textbf{Hyper-parameter} & \textbf{Search   range}\\\hline   
Depth                                   & {[}2,3{]}         \\
Split criterion                         & ‘gini’, ‘entropy’ \\
Number of features to use at each split & {[}10\%,100\%{]} 
\\ \hline
\end{tabular}
\label{tab:hyper}
\end{table}

\begin{table}[!t]
\centering
\caption{\footnotesize{List of MOR features: Interval duration.}}
\begin{tabular}{p{60pt}|p{155pt}}\hline
\textbf{Feature} & \textbf{Definition}   \\ \hline 
DmedQRS          & Median P-wave [ms] \\
DmeanQRS         & Mean P-wave [ms] \\
DstdQRS          & Standard deviation P-wave [ms] \\
DmaxQRS          & Maximum P-wave [ms] \\ \hline
DmedPR           & Median PR interval [ms] \\
DmeanPR          & Mean PR interval [ms] \\
DstdPR           & Standard deviation PR interval [ms] \\
DmaxPR           & Max PR interval [ms] \\ \hline
DmedPRseg        & Median PR segment [ms] \\
DmeanPRseg       & Mean PR segment [ms] \\
DstdPRseg        & Standard deviation PR segment [ms] \\
DmaxPRseg        & Max PR segment [ms] \\ \hline
DmedQRS          & Median QRS interval [ms] \\
DmeanQRS         & Mean QRS interval [ms] \\
DstdQRS          & Standard deviation QRS interval [ms] \\
DmaxQRS          & Maximum QRS interval [ms] \\ \hline
DmedQT           & Median QT interval [ms] \\
DmeanQT          & Mean QT interval [ms] \\
DstdQT           & Standard deviation QT interval [ms] \\
DmaxQT           & Maximum QT interval [ms] \\ \hline
DmedT            & Median T-wave [ms] \\
DmeanT           & Mean T-wave [ms] \\
DstdT            & Standard deviation T-wave [ms] \\
DmaxT            & Maximum T-wave [ms] \\ \hline
DmedRR           & Median RR interval [ms] \\
DmeanRR          & Mean RR interval [ms] \\
DstdRR           & Standard deviation RR interval[ms] \\
DmaxRR           & Maximum RR interval [ms] \\ \hline
IRmed            & Median of the RR interval ratio [n.u.] \\
IRmean           & Mean RR interval ratio [n.u.] \\
IRstd            & Standard deviation RR interval ratio [n.u.] \\
IRmax            & Maximum RR interval ratio [n.u.] \\ \hline
MDPR             & Median of the PR interval duration as defined by Mao et al. \cite{mao2019automated} [ms] \\
MAPR             & Mean of the PR interval duration as defined by Mao et al. \cite{mao2019automated} [ms] \\ \hline
DmedQT\_b        & Corrected QT interval (QTc) by Bazett [ms] \\
DmedQT\_fre      & Corrected QT interval (QTc) by Fridericia [ms] \\
DmedQT\_fra      & Corrected QT interval (QTc) by Framingham [ms] \\
DmedQT\_hod      & Corrected QT interval (QTc) by Hodges [ms] \\\hline                                      
\end{tabular}
\label{tab:MOR1}
\end{table}

\begin{table}[!t]
\centering
\caption{\footnotesize{List of MOR features: Waves characteristics.}}
\begin{tabular}{p{60pt}|p{155pt}}\hline
\textbf{Feature} & \textbf{Definition}   \\\hline
medP         & Median P-wave amplitude [1e-4v] \\
meanP        & Mean P-wave amplitude [1e-4v] \\
stdP         & Standard deviation P-wave amplitude [1e-4v] \\
maxP         & Maximum P-wave amplitude [1e-4v] \\ \hline
medST        & Median ST segment amplitude [1e-4v] \\
meanST       & Mean ST segment amplitude [1e-4v] \\
stdST        & Standard deviation ST segment amplitude [1e-4v] \\
maxST        & Maximum ST segment amplitude [1e-4v] \\ \hline
medR         & Median R-wave amplitude [1e-4v] \\
meanR        & Mean R-wave amplitude [1e-4v] \\
stdR         & Standard deviation R-wave amplitude [1e-4v] \\
maxR         & Maximum R-wave amplitude [1e-4v] \\ \hline
medDR        & Median R-wave amplitude variation [1e-4v] \\
meanDR       & Mean R-wave amplitude variation [1e-4v] \\
stdDR        & Standard R-wave amplitude variation [1e-4v] \\
maxDR        & Maximum R-wave amplitude variation [1e-4v] \\ \hline
medQRS       & Median QRS segment amplitude [1e-4v] \\
meanQRS      & Mean QRS segment amplitude [1e-4v] \\
stdQRS       & Standard deviation QRS segment amplitude [1e-4v] \\
maxQRS       & Maximum QRS segment amplitude [1e-4v] \\ \hline
SmedQRS      & Median QRS interval area [1e-4v] \\
SmeanQRS     & Mean QRS interval area [1e-4v] \\
SstdQRS      & Standard deviation QRS interval area [1e-4v] \\
SmaxQRS      & Maximum QRS interval area [1e-4v] \\ \hline
SmedQRSdiff  & Median of the difference between the area of the tested QRS and a reference QRS [\%] \\
SmeanQRSdiff & Mean of the difference between the area of the tested QRS and a reference QRS [\%] \\
SstdQRSdiff  & Standard deviation of the difference between the area of the tested QRS and a reference QRS [\%] \\
SmaxQRSdiff  & Maximum of the difference between the area of the tested QRS and a reference QRS [\%] \\ \hline
DmedQRSdiff  & Median of the difference between the width of the tested QRS and a reference QRS [\%] \\
DmeanQRSdiff & Mean of the difference between the width of the tested QRS and a reference QRS [\%] \\
DstdQRSdiff  & Standard deviation of the difference between the width of the tested QRS and a reference QRS [\%] \\
DmaxQRSdiff  & Maximum of the difference between the width of the tested QRS and a reference QRS [\%] \\ \hline
medJ         & Median J-point amplitude [1e-4v] \\
meanJ        & Mean J-point amplitude  [1e-4v] \\
stdJ         & Standard deviation J-point amplitude [1e-4v] \\
maxJ         & Maximum J-point amplitude [1e-4v] \\ \hline
\end{tabular}
\label{tab:MOR2}
\end{table}

\end{document}